\documentclass{article} 
\usepackage{iclr2027_conference,times}

\usepackage{amsmath,amsfonts,bm}

\def\eqref#1{equation~\ref{#1}}

\def\1{\bm{1}}

\DeclareMathAlphabet{\mathsfit}{\encodingdefault}{\sfdefault}{m}{sl}
\SetMathAlphabet{\mathsfit}{bold}{\encodingdefault}{\sfdefault}{bx}{n}

\usepackage{hyperref}
\usepackage{url}
\usepackage{graphicx}
\usepackage{multirow}
\usepackage{booktabs}
\usepackage[table]{xcolor}

\title{ReEnvision: Future-Guided Selective Action Repair for Long-Horizon Manipulation}

\author{
\textbf{Xiyin Zeng}$^{1,2*}$ \quad
\textbf{Xuewei Huang}$^{1,2*}$ \quad
\textbf{Yi Lu}$^{1}$ \quad
\textbf{Weihao Yuan}$^{2,3\dagger}$ \quad
\textbf{Hao Wang}$^{1\dagger}$ \\
$^1$ Information Hub, The Hong Kong University of Science and Technology (Guangzhou) \\
$^2$ Daimon Robotics \\
$^3$ School of Intelligence Science and Technology, Nanjing University
}

\iclrfinalcopy 
\begin{document}

\maketitle

\begin{abstract}

Long-horizon robotic manipulation requires robust execution over sequences of interdependent actions. Existing methods typically handle execution deviations by replanning the remaining action chunk or correcting actions step by step as new observations arrive. However, a local deviation may affect only a few actions under distribution shift, yet replanning regenerates the entire chunk from an unfamiliar state, potentially introducing errors into otherwise valid actions. Step-wise correction may react too late, as it does not explicitly anticipate how the deviation affects subsequent actions.
We propose ReEnvision, a future-guided selective action repair framework that updates a joint future-action belief using execution feedback. It then learns how the resulting belief revision affects each unexecuted action. The resulting belief revision enables selective repair of only the affected actions within a cached action chunk, preserving valid actions and reserving full replanning for severe deviations.
On CALVIN, L-CALVIN, and RLBench, ReEnvision achieves 86.6\% 5-step success, improves 10-step success from 56\% to 63\%, and reaches an average success rate of 88.8\%, respectively. On five real-world manipulation tasks, it achieves an average success rate of 82.8\%. Controlled repair experiments further demonstrate the benefits of localizing affected actions and preserving valid ones during execution. Our project page is available at https://reenvision.pages.dev.

\end{abstract}

\section{Introduction}

Long-horizon robotic manipulation requires robots to coordinate multiple dependent actions over an extended execution horizon, while preserving consistency across stages and recovering from intermediate deviations~\citep{zhou2024spire,ZhouP-RSS-26}. Such capability is important for real-world manipulation, where many tasks involve sustained, multi-stage interactions rather than isolated actions~\citep{NEURIPS2024_ca92ff06, chai2026evolvenavproactivepreflectionselfevolving}. However, reliable long-horizon execution remains challenging: an error at one stage can alter the state encountered by subsequent actions, causing later actions to become inappropriate and allowing errors to accumulate until the overall task fails~\citep{Tan_2026_CVPR}.

Long-horizon manipulation has been approached from multiple perspectives, including hierarchical planning, predictive world modeling, and adaptive execution~\citep{chai2025causalmacecausalityempoweredmultiagents, Ma_2026_CVPR}. Hierarchical planning and control methods decompose long tasks into subtasks, while world models anticipate future states and use them to guide actions~\citep{ICLR2026_9da515b1, LiL-RSS-26}. More recently, adaptive execution methods incorporate new observations during execution to adjust planned behavior: Future Forward Dynamics Causal Attention (FFDC)~\citep{wang2026trustimaginationadaptiveaction} verifies whether the current rollout should continue or be replanned, while Asynchronous Action Chunk Correction (A2C2)~\citep{sendai2025leaveobservationbehindrealtime} uses each newly observed frame to predict a residual update to the actions being executed. In contrast, Real-Time Robot Execution with Masked Action Chunking (REMAC)~\citep{ICLR2026_c3583444} preserves actions that are already committed and resamples only the remaining portion of the chunk. 
Despite these advances, existing methods primarily address execution deviations through rollout-level replanning, step-wise correction, or resampling the editable portion of an action chunk. They do not explicitly estimate which unexecuted actions are affected by an observed deviation. Under out-of-distribution (OOD) deviations, regenerating the entire remaining chunk from an unfamiliar intermediate state may disrupt established alignment and contact, amplifying local errors into downstream failures~\citep{Chen_2026_CVPR}. Our analysis identifies pronounced replanning failures in approximately 11\% of OOD sequences, with 55\% occurring in direction-sensitive manipulation and 36\% in contact- or pose-sensitive tasks. Meanwhile, step-wise correction may fail to anticipate how current deviations affect subsequent actions.

\begin{figure}[t]
    \centering
    \includegraphics[width=\linewidth]{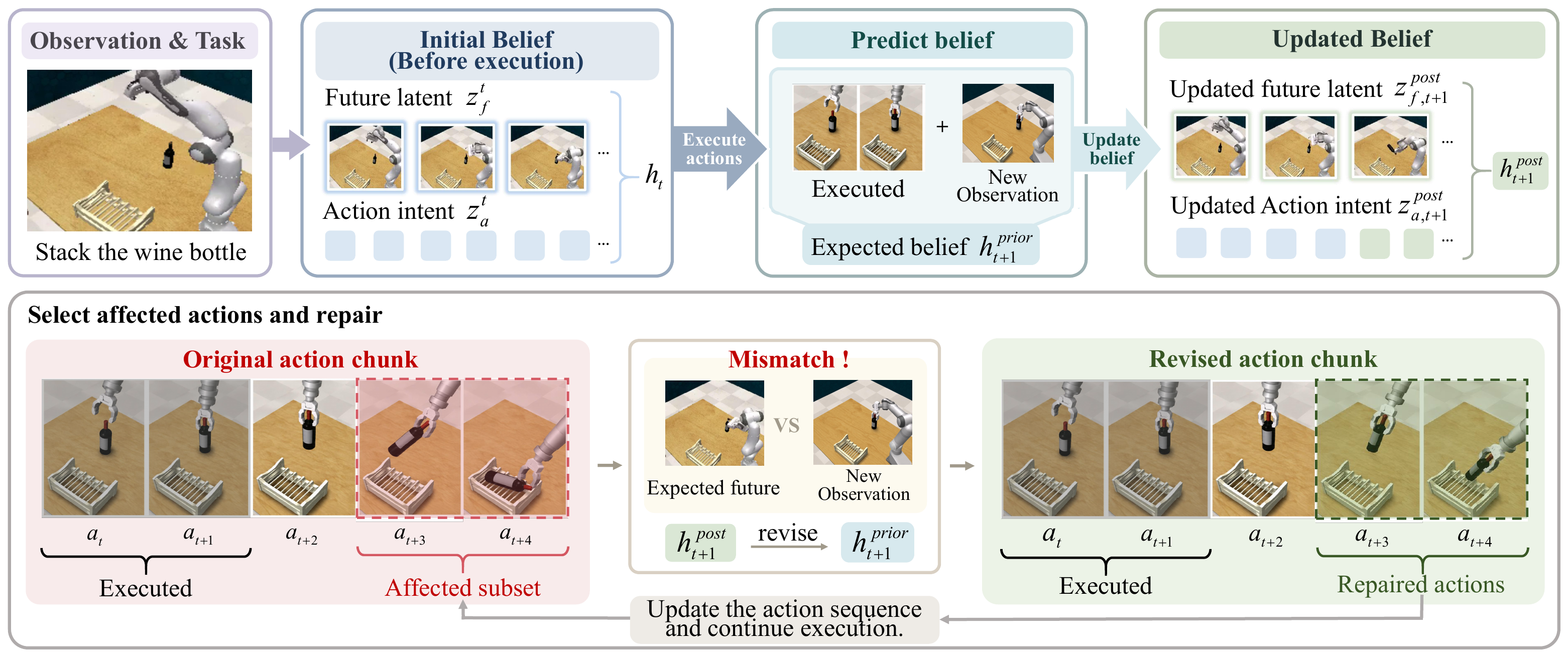}
    \caption{
    Overview of ReEnvision. ReEnvision maintains a joint future-action belief \(h_t\) and revises it with new observations. The revision identifies affected actions in the cached chunk, enabling selective repair. Unlike methods that rely on full replanning, ReEnvision repairs only affected actions within the current chunk while preserving unaffected ones.
    }
    \label{fig:overview}
    \vspace{-2em}
\end{figure}

Addressing this limitation requires reasoning about the expected future and planned actions together: execution errors change what happens next, and these changes can in turn make some planned actions no longer suitable~\citep{ye2026worldactionmodelszeroshot}. By jointly modeling these two signals, the robot can reason about how an observed deviation changes what should happen next and how that change affects subsequent actions. This makes it possible to determine whether the remaining action chunk requires a global update or only a local adjustment. When only part of the chunk is affected, selectively repairing those actions preserves unaffected ones without regenerating the entire sequence.

Building on this idea, we propose ReEnvision, a future-guided selective action repair framework that revises a joint future-action belief and anticipates how the revision affects individual actions within a cached chunk as illustrated in Fig.~\ref{fig:overview}. Given the current observation and task, a pretrained future generator predicts visual evolution, which is combined with the observed state to infer the corresponding action intent.
 Together, they form a joint future-action belief that conditions a diffusion policy to generate a multi-step action chunk.
As new observations arrive, a latent dynamics model predicts how the joint belief should evolve following the executed action and updates the prediction using the observed outcome. The difference between the predicted and updated beliefs defines the belief revision. 
A temporal impact model then estimates how this belief revision affects each remaining action in terms of correction, validity, and uncertainty. These predictions enable selective repair of the affected actions while preserving unaffected ones, with full replanning reserved for broadly affected or highly uncertain suffixes. In experiments, ReEnvision achieves relative improvements of 2.5\%, 12.5\%, and 2.5\% over the strongest reported baselines on CALVIN~\citep{mees2022calvinbenchmarklanguageconditionedpolicy} 5-step success, L-CALVIN~\citep{fan2025long} 10-step success, and real-world manipulation, respectively, while achieving 88.8\% average success on RLBench~\citep{james2019rlbenchrobotlearningbenchmark}.

Our contributions are summarized as follows:

(i) We find that changes in the expected future typically affect only a local subset of actions within the remaining chunk; selectively repairing these actions is more effective than rewriting the entire chunk and enables correction before they are executed.

(ii) We propose ReEnvision, which learns how revisions in a joint future-action belief affect each unexecuted action and uses the predicted action-level impact to determine the scope of intervention, preserving unaffected actions while selectively repairing affected segments.

(iii) Extensive experiments demonstrate that ReEnvision achieves 86.6\% 5-step success on CALVIN, 88.8\% average success on RLBench, and 82.8\% average success on five real-world manipulation tasks, with a 12.5\% relative improvement in 10-step success on L-CALVIN.

\section{Related work}
\subsection{Hierarchical Planning and Control}
Hierarchical planning and control methods decompose long-horizon manipulation into subgoals or execution phases, thereby reducing the effective decision horizon and improving temporal consistency~\citep{Zhao-RSS-23}. Hierarchical Action Models for Open-World Robot Manipulation (HAMSTER)~\citep{ICLR2025_3bfee3bc} first predicts a coarse end-effector trajectory, then conditions a low-level policy on this trajectory to generate precise 3D actions. Similarly, Long-VLA~\citep{fan2025long} instead decomposes each subtask into moving and interaction phases, and generates actions conditioned on the current phase to improve transitions between subtasks. Going beyond static decomposition, Reflective Contrastive Alignment and Planning Architecture (ReCAPA)~\citep{zeng2026recapahierarchicalpredictivecorrection} further introduces predictive correction across action, subgoal, and trajectory levels, allowing local deviations to be corrected before they propagate into cascading failures. Despite these advances, such methods primarily operate over action-centric representations, without explicitly modeling how the physical world should evolve toward the task goal. 

\subsection{Predictive World Models}
Predictive world models forecast how the environment will evolve under candidate actions and use these predicted futures to guide subsequent control~\citep{feng2025reflective}. Latent Diffusion World Model (LaDi-WM)~\citep{huang2025ladi} uses a diffusion-based world model to predict future states in a learned latent space, then conditions a diffusion policy on these predictions to refine actions. More recently, Mixture of Latent Actions (MoLA)~\citep{li2026imaginedfuturesexecutableactions} focuses on grounding imagined futures into executable control, converting future visual transitions into latent actions through inverse dynamics before decoding them into robot commands. In contrast, Decoupled Visual Forward and Inverse Dynamics Pretraining (DeFI)~\citep{ICLR2026_793bfa8f} decouples the two learning objectives, separately training forward dynamics to predict futures and inverse dynamics to infer actions. However, these methods typically use predicted futures to guide action generation, so execution deviations may render the remaining actions inconsistent with the updated future~\citep{gao2026dreamdojogeneralistrobotworld}.

\subsection{Adaptive Action Execution}
Adaptive action execution mitigates the mismatch between the updated future and the remaining committed actions by using execution feedback for online correction or replanning~\citep{liang2026adaptiveactionchunkinginferencetime, yao2026worldmodels}. FFDC ~\citep{wang2026trustimaginationadaptiveaction} compares predicted visual futures and actions with real observations to decide whether to continue the current rollout or replan, while A2C2~\citep{sendai2025leaveobservationbehindrealtime} uses visual feedback at each step to predict a residual correction to the currently executed chunk action. More selectively, REMAC~\citep{ICLR2026_c3583444} preserves already committed actions while resampling the remaining editable portion of the action chunk to maintain consistency during execution.
These methods improve adaptive execution through rollout-level replanning decisions, step-wise residual correction, or masked action generation. However, they do not explicitly characterize how changes in the expected future affect individual unexecuted actions within a cached chunk~\citep{Bi_2026_CVPR}. 
Consequently, these corrections may unnecessarily modify unaffected actions or fail to anticipate downstream effects on subsequent actions.

\section{Methodology}
\subsection{Framework Overview}
\begin{figure}[t]
    \centering
    \includegraphics[width=\linewidth]{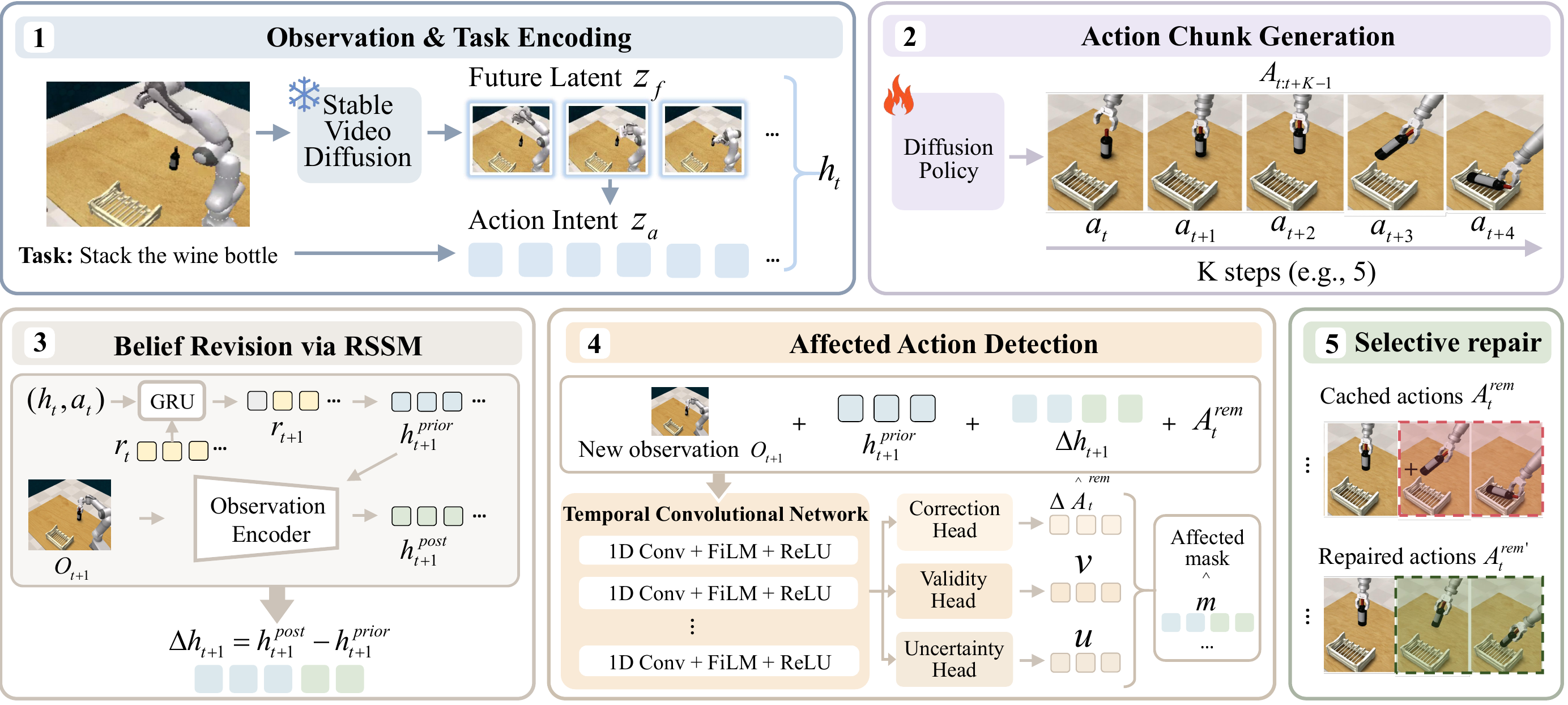}
    \caption{
    In ReEnvision, joint future-action belief \(h_t\) conditions action-chunk generation and is revised after each new observation. The resulting belief revision is used to predict per-action correction, validity, and uncertainty, enabling selective repair of the affected cached actions.
    }
    \label{fig:architecture}
    \vspace{-1em}
\end{figure}

To avoid both broad replanning from unfamiliar intermediate states and delayed step-wise correction, execution feedback should be used to anticipate how the expected future has changed and where its effects will appear in the remaining action chunk. This motivates localizing the affected actions in advance and revising only that portion while preserving the rest. To this end, we propose ReEnvision as shown in Fig.~\ref{fig:architecture}, which uses Stable Video Diffusion (SVD)~\citep{blattmann2023stablevideodiffusionscaling} to predict visual evolution from the current observation and task instruction. Conditioned on the current state and anticipated future, an intent generator produces an action representation. Together with the future latent, it forms a joint belief that conditions a diffusion-based policy to generate an action chunk. As new observations arrive, a Recurrent State-Space Model (RSSM)~\citep{hafner2019learning} updates the joint belief and measures its deviation from the expected belief transition. A Temporal Convolutional Network (TCN)~\citep{bai2018empirical} then uses this belief revision to estimate how each action in the remaining chunk is affected, leading to keep, selective correction, or replanning.

\subsection{Joint Future-Action Belief}

Given the current observation \(o_t\) and language instruction \(l\), a pretrained SVD model extracts a base future representation \(z_f^t \equiv z_f^{\mathrm{base}}\) of the expected task-relevant evolution. During training, a target future representation \(z_f^{\mathrm{target}}\) is constructed from successful trajectories to represent the desired execution stage. A Multilayer Perceptron (MLP)-based action-intent generator predicts an action-intent representation \(z_a^t\), conditioned on the current observation, the base and target future representations, their difference \(\Delta z_f=z_f^{\mathrm{target}}-z_f^t\), and the task instruction. 
The generator is trained using an \(L_1\) loss against action-intent targets derived from successful action chunks. During training, positive, identity, and negated revision pairs are sampled equally, with identity pairs mitigating the training–inference mismatch under zero revision.

At inference, the initial action intent is generated using a zero future revision, with \(z_f^{\mathrm{target}}=z_f^t\) and \(\Delta z_f=0\), without requiring a ground-truth target future. The future and action-intent representations are combined into a joint future-action belief,
\begin{equation}
h_t = (z_f^t,z_a^t),
\label{eq:joint_belief}
\end{equation}
which captures both the expected future and the corresponding action intent.

To generate executable actions, \(z_a^t\) is introduced as an additional conditioning signal to the pretrained diffusion policy. The original conditioning \((o_t,l,z_f^t)\) is extended to \((o_t,l,z_f^t,z_a^t)\), while the training objective remains the original diffusion action-prediction objective on ground-truth action chunks. The resulting policy produces a \(K\)-step action chunk \(A_t=[a_t,\ldots,a_{t+K-1}]\).

\subsection{Belief Revision}

After generating an action chunk, the robot executes \(a_t\) and observes \(o_{t+1}\) as feedback on the resulting environment transition. Since the belief naturally evolves during execution, a change from \(h_t\) to the next belief does not by itself indicate an execution deviation. 
An RSSM therefore models the expected belief transition and incorporates execution feedback to capture discrepancies between the predicted and observation-updated beliefs. Given the current belief \(h_t\) and executed action \(a_t\), the prior branch predicts the expected next belief \(h_{t+1}^{\mathrm{prior}}\). The posterior branch additionally incorporates the new observation \(o_{t+1}\) to infer the updated belief \(h_{t+1}^{\mathrm{post}}\).

During training, paired future--action latent representations are collected from consecutive execution steps of the same rollout. The current belief \(h_t\) and target belief \(h_{\mathrm{target}}\) correspond to the current and next execution steps, respectively. Their difference defines the teacher latent displacement,
\begin{equation}
\Delta h^* = h_{\mathrm{target}}-h_t.
\label{eq:teacher_displacement}
\end{equation}
Both RSSM branches are supervised to predict this displacement:
\begin{equation}
\mathcal{L}_{\mathrm{RSSM}} = \ell(h_{t+1}^{\mathrm{prior}}-h_t,\Delta h^*) + \ell(h_{t+1}^{\mathrm{post}}-h_t,\Delta h^*),
\label{eq:rssm_loss}
\end{equation}
where \(\ell\) denotes the Smooth \(L_1\) loss. The prior learns the expected belief transition without access to the new observation, while the posterior incorporates the observed outcome to refine the predicted update. During execution, the belief revision is defined as the difference between the posterior and prior beliefs, \(\Delta h_{t+1}=h_{t+1}^{\mathrm{post}}-h_{t+1}^{\mathrm{prior}}\). This revision captures the observation-induced update to the expected belief transition and guides selective action repair. The posterior belief is then maintained for subsequent execution, \(h_{t+1}\leftarrow h_{t+1}^{\mathrm{post}}\).

\subsection{Selective Action Repair}

\subsubsection{Belief-Guided Suffix Repair}

After executing \(a_t\), the remaining actions are cached as \(A_t^{\mathrm{rem}}=[a_{t+1},\ldots,a_{t+K-1}]\). Since the belief revision \(\Delta h_{t+1}\) may affect only part of this suffix, a TCN \(S_\psi\) predicts an action correction \(\Delta\hat{a}_k\), validity score \(v_k\), and uncertainty \(u_k\) for each remaining action:
\begin{equation}
\left(\Delta\hat{A}_t^{\mathrm{rem}},\mathbf{v},\mathbf{u}\right)
=
S_\psi\left(o_{t+1},h_{t+1}^{\mathrm{prior}},A_t^{\mathrm{rem}},\Delta h_{t+1}\right),
\label{eq:sensitivity_tcn}
\end{equation}
where \(k=1,\ldots,K-1\) indexes the cached actions, larger \(v_k\) indicates higher action validity, and larger \(u_k\) indicates greater uncertainty in determining whether the action requires correction.

The affected actions are selected by optimizing a binary mask \(\mathbf{m}\in\{0,1\}^{K-1}\):
\begin{equation}
\begin{aligned}
\hat{\mathbf{m}}=\arg\min_{\mathbf{m}}\quad
&\lambda_v\sum_{k=1}^{K-1}(1-m_k)(1-v_k)
+\lambda_m\sum_{k=1}^{K-1}m_k\\
&+\lambda_c\sum_{k=1}^{K-2}|m_{k+1}-m_k|
+\lambda_u\sum_{k=1}^{K-1}m_ku_k.
\end{aligned}
\label{eq:mask_optimization}
\end{equation}
The objective penalizes leaving low-validity actions unrepaired, modifying unnecessary actions, fragmented repair regions, and repairing highly uncertain actions. The corresponding weights are set to \(\lambda_v=1\), \(\lambda_m=0.5\), \(\lambda_c=0.15\), and \(\lambda_u=0.15\), favoring compact and reliable repair regions.

The optimized mask determines whether the cached suffix should be kept, selectively repaired, or discarded. The repair ratio \(r=\frac{1}{K-1}\sum_{k=1}^{K-1}\hat{m}_k\) measures the fraction of affected actions, \(s\) denotes the number of disconnected affected segments, and \(u_{\max}=\max_{1\leq k\leq K-1}u_k\) measures the maximum uncertainty across the entire suffix. Full replanning is triggered when \(u_{\max}\geq\eta\). Otherwise, the suffix is retained when no action is selected, and selective repair is applied when \(r\leq\rho\) and \(s\leq s_{\max}\). If either condition is violated, the cached suffix is discarded and fully replanned. Here, \(\rho\), \(s_{\max}\), and \(\eta\) control the maximum repair extent, fragmentation, and uncertainty, respectively.

For selective repair, only the selected actions are updated:
\begin{equation}
A_t^{\mathrm{rem}\prime}
=
A_t^{\mathrm{rem}}
+
\hat{\mathbf{m}}\odot\Delta\hat{A}_t^{\mathrm{rem}},
\label{eq:selective_patch}
\end{equation}
with the correction magnitude clipped to a bounded range and smoothing applied within the selected repair region to maintain trajectory continuity. Under full replanning, the cached suffix is discarded, and a new action chunk is generated by the diffusion policy conditioned on the posterior belief \(h_{t+1}^{\mathrm{post}}\).

\subsubsection{Action Impact Model Training}

\(S_\psi\) is trained on outcome-guided action pairs to approximate belief-induced corrections. Each pair consists of lower- and higher-return action chunks, \(A^{-}\) and \(A^{+}\), generated by the same policy under matched observation, task, diffusion noise, and cache conditions.
\begin{equation}
A^{-}=A_{\mathrm{neg}}, \qquad
A^{+}=A_{\mathrm{pos}}.
\label{eq:paired_counterfactual}
\end{equation}

Both \(A^{-}\) and \(A^{+}\) are full action chunks generated by the same frozen policy. Their first \(K-1\) actions, denoted by \(A_{\mathrm{rem}}^{-}\) and \(A_{\mathrm{rem}}^{+}\), are temporally aligned with the cached suffix \(A_t^{\mathrm{rem}}\). Their difference, \(\Delta A_{\mathrm{rem}}^{*}=A_{\mathrm{rem}}^{+}-A_{\mathrm{rem}}^{-}\), provides an outcome-guided correction from lower- to higher-return actions. At inference, pre-revision actions are retrieved directly from the cached suffix \(A_t^{\mathrm{rem}}\). 
For each remaining action \(k\), the validity target indicates whether its outcome-guided correction magnitude remains within a predefined threshold:

\begin{equation}
v_k^{*}=\mathbb{I}\left[\|\Delta a_k^{*}\|_2\leq\tau\right],
\label{eq:validity_label}
\end{equation}
where \(k=1,\ldots,K-1\) indexes the cached suffix, and 
\(v_k^*=1\) indicates that the outcome-guided correction magnitude does not exceed the threshold \(\tau\).
An uncertainty target \(u_k^{*}\) is further constructed from the distance between \(\|\Delta a_k^{*}\|_2\) and \(\tau\), assigning higher uncertainty to actions closer to the validity decision boundary and lower uncertainty to those farther away. 
The threshold \(\tau\) is used only to construct offline supervision targets. During inference, predicted validity and uncertainty guide structured mask selection without directly thresholding predicted corrections.

\(S_\psi\) is jointly trained to predict outcome-guided action corrections, validity, and uncertainty:
\begin{equation}
\mathcal{L}_{\mathrm{TCN}}=\left\|\Delta\hat{A}_t^{\mathrm{rem}}-\Delta A_{\mathrm{rem}}^{*}\right\|_1+\lambda_{\mathrm{val}}\operatorname{BCE}\left(\mathbf{v},\mathbf{v}^{*}\right)+\lambda_{\mathrm{unc}}\left\|\mathbf{u}-\mathbf{u}^{*}\right\|_1.
\label{eq:tcn_loss}
\end{equation}
The loss weights are set to \(\lambda_{\mathrm{val}}=1.0\) and \(\lambda_{\mathrm{unc}}=0.1\). This outcome-guided supervision is used to approximate the action changes induced by belief revision. During execution, predicted corrections are applied only to the selected actions in Eq.~\ref{eq:selective_patch}, while unaffected actions are preserved.

\begin{table}[t]
\caption{CALVIN~\citep{mees2022calvinbenchmarklanguageconditionedpolicy} ABC$\rightarrow$D results. We report the average success rate over 1,000 rollouts and the average number of consecutively completed tasks (Avg. Len.). ReEnvision demonstrates clear improvements over the baselines, particularly on longer-horizon sequences. }
\centering
\small
\setlength{\tabcolsep}{5pt}
\begin{tabular}{lcccccc}
\toprule
\textbf{Method}
& \textbf{1}
& \textbf{2}
& \textbf{3}
& \textbf{4}
& \textbf{5}
& \textbf{Avg. Len.} $\uparrow$ \\
\midrule

$\pi_{0.5}^*$~\citep{intelligence2025pi05visionlanguageactionmodelopenworld}
& 94.8 & 87.4 & 78.2 & 71.7 & 64.3 & 3.97 \\

VPP~\citep{hu2025videopredictionpolicygeneralist}
& 96.5 & 90.9 & 86.6 & 82.0 & 76.9 & 4.33 \\

REMAC~\citep{ICLR2026_c3583444} 
& 96.0 & 91.4 & 88.3 & 84.9 & 78.4 & 4.39 \\

PALM~\citep{liu2026palmprogressawarepolicylearning}
& 96.9 & 93.8 & 89.3 & 85.9 & 82.0 & 4.48 \\

DeFI~\citep{ICLR2026_793bfa8f}
& 97.9 & 94.2 & 90.7 & 87.0 & 81.2 & 4.51 \\

VITA~\citep{Ma_2026_CVPR}
& \textbf{99.1} & \textbf{94.9} & 91.2 & 87.8 & 84.5 & 4.57 \\

\textbf{ReEnvision}
& 97.8
& 94.5
& \textbf{93.0}
& \textbf{90.1}
& \textbf{86.6}
& \textbf{4.62} \\

\bottomrule
\end{tabular}
\label{tab:calvin_results}
\end{table}

\subsection{Training and Inference}

\subsubsection{Training}

The framework is trained in stages. Future representations from the pretrained SVD are first used to train the action-intent generator, after which the diffusion policy is further trained to generate action chunks conditioned on both future and action-intent representations. Rollouts from the resulting policy are then used to train the RSSM, with both prior and posterior branches supervised by the teacher latent displacement \(\Delta h^*=h_{\mathrm{target}}-h_t\). Finally, matched lower- and higher-return action chunks from the frozen diffusion policy provide outcome-guided correction targets for training \(S_\psi\) to predict per-action corrections, validity, and uncertainty.

\subsubsection{Inference}

At initialization, the current observation and task instruction are used to construct the initial joint belief and generate an action chunk. After each action is executed, the RSSM predicts the next belief and incorporates the new observation to obtain the posterior belief. The pre-revision actions are retrieved directly from the cached suffix \(A_t^{\mathrm{rem}}\), without regeneration. Given the belief revision \(\Delta h_{t+1}\) and cached suffix, \(S_\psi\) predicts per-action corrections, validity, and uncertainty. The suffix is retained when no action is selected and uncertainty is low, and selectively repaired when the affected region is sufficiently localized. Full replanning is triggered when the affected region is too large, fragmented, or uncertain, generating a new action chunk conditioned on \(h_{t+1}^{\mathrm{post}}\).

\begin{table}[t]
\caption{Comparison with prior methods on the L-CALVIN~\citep{fan2025long}, using 100 evaluation sequences of 10 consecutive instructions under the ABC$\rightarrow$D setting.}
\label{tab:lcalvin_results}
\centering
\small
\setlength{\tabcolsep}{3.5pt}
\begin{tabular}{lccccccccccc}
\toprule
\textbf{Method}
& \multicolumn{10}{c}{\textbf{Tasks Completed in Sequence}}
& \textbf{Avg. Len.} \\
\cmidrule(lr){2-11}
& \textbf{1} & \textbf{2} & \textbf{3} & \textbf{4} & \textbf{5}
& \textbf{6} & \textbf{7} & \textbf{8} & \textbf{9} & \textbf{10}
& \\
\midrule

VLAS~\citep{zhao2025vlasvisionlanguageactionmodelspeech}
& 0.88 & 0.80 & 0.72 & 0.58 & 0.49
& 0.46 & 0.34 & 0.28 & 0.19 & 0.16
& 4.90 \\

RoboVLMs~\citep{li2026mattersbuildingvisionlanguageactionmodels}
& 0.90 & 0.80 & 0.78 & 0.70 & 0.64
& 0.60 & 0.51 & 0.41 & 0.36 & 0.34
& 6.04 \\

REMAC~\citep{ICLR2026_c3583444} 
& 0.94 & 0.90 & 0.82 & 0.74 & 0.71
& 0.58 & 0.54 & 0.50 & 0.47 & 0.41
& 6.61 \\

DeFI~\citep{ICLR2026_793bfa8f}
& 0.98 & 0.96 & 0.92 & 0.86 & 0.75
& 0.71 & 0.59 & 0.57 & 0.52 & 0.44
& 7.30 \\

Long-VLA~\citep{fan2025long}
& \textbf{1.00} & \textbf{1.00} & \textbf{0.98} & 0.91 & 0.85
& 0.82 & \textbf{0.79} & 0.70 & 0.63 & 0.56
& 8.24 \\

\textbf{ReEnvision}
& 0.99 & 0.98 & 0.95 & \textbf{0.92} & \textbf{0.89}
& \textbf{0.83} & 0.77 & \textbf{0.72} & \textbf{0.68} & \textbf{0.63}
& \textbf{8.36} \\

\bottomrule
\end{tabular}
\end{table}

\begin{table*}[t]
\caption{
Performance comparison on 18 RLBench~\citep{james2019rlbenchrobotlearningbenchmark} manipulation tasks, with 50 evaluation episodes per task. We report task success rates (\%) and the average success rate (Avg. SR) across all 900 episodes. Best results are shown in bold.
}
\label{tab:rlbench_results}
\centering
\small
\setlength{\tabcolsep}{2.4pt}

\begin{tabular}{lccccccccc}
\toprule
\textbf{Method}
& \textbf{Avg. SR}
& \shortstack{\textbf{Close}\\\textbf{Jar}}
& \shortstack{\textbf{Insert}\\\textbf{Peg}}
& \shortstack{\textbf{Light}\\\textbf{Bulb}}
& \shortstack{\textbf{Meat off}\\\textbf{Grill}}
& \shortstack{\textbf{Open}\\\textbf{Drawer}}
& \shortstack{\textbf{Place}\\\textbf{Cups}}
& \shortstack{\textbf{Shape}\\\textbf{Sorter}}
& \shortstack{\textbf{Place}\\\textbf{Wine}} \\
\midrule

HDFlow~\citep{gireesh2026hdflowhierarchicaldiffusionflowplanning}
& 87.5
& 96.0 & 93.3 & 88.7 & 98.7
& 82.0 & 56.7 & \textbf{73.3} & 94.7 \\

TVVE~\citep{bai2026learningacttaskawarevirtual}
& 86.6
& \textbf{100.0} & \textbf{98.0} & 86.0 & 94.0
& 90.0 & 54.0 & 62.0 & 92.0 \\

SpatialActor~\citep{shi2026spatialactorexploringdisentangledspatial}
& 87.4
& 94.0 & 93.3 & 88.7 & 98.7
& 82.0 & 56.7 & \textbf{73.3} & 94.7 \\

BridgeVLA~\citep{li2025bridgevlainputoutputalignmentefficient}
& 85.7
& 98.0 & 80.0 & 86.0 & \textbf{100.0}
& \textbf{98.0} & 48.0 & 38.0 & 90.0 \\

\textbf{ReEnvision}
& \textbf{88.8}
& 98.0 & 82.0 & \textbf{90.0} & \textbf{100.0}
& 90.0 & 60.0 & 50.0 & \textbf{96.0} \\

\bottomrule
\end{tabular}

\vspace{1.5mm}

\begin{tabular}{lcccccccccc}
\toprule
\textbf{Method}
& \shortstack{\textbf{Push}\\\textbf{Buttons}}
& \shortstack{\textbf{Put}\\\textbf{Groceries}}
& \shortstack{\textbf{Put in}\\\textbf{Drawer}}
& \shortstack{\textbf{Put in}\\\textbf{Safe}}
& \shortstack{\textbf{Reach \&}\\\textbf{Drag}}
& \shortstack{\textbf{Slide}\\\textbf{Block}}
& \shortstack{\textbf{Stack}\\\textbf{Blocks}}
& \shortstack{\textbf{Stack}\\\textbf{Cups}}
& \shortstack{\textbf{Sweep}\\\textbf{Dustpan}}
& \shortstack{\textbf{Turn}\\\textbf{Tap}} \\
\midrule

HDFlow
& \textbf{100.0} & 72.0 & 98.7 & 96.7 & \textbf{100.0}
& 91.3 & 56.0 & 81.3 & \textbf{100.0} & 95.3 \\

TVVE
& \textbf{100.0}
& 74.0 & \textbf{100.0} & 78.0 & \textbf{100.0}
& \textbf{100.0} & 74.0 & 64.0 & 92.0 & \textbf{100.0} \\

SpatialActor
& \textbf{100.0}
& 72.0 & 98.7 & 96.7 & \textbf{100.0}
& 91.3 & 56.0 & 81.3 & \textbf{100.0} & 95.3 \\

BridgeVLA
& \textbf{100.0}
& 80.0 & 88.0 & \textbf{100.0} & \textbf{100.0}
& 98.0 & 76.0 & 80.0 & 90.0 & 92.0 \\

\textbf{ReEnvision}
& 90.0
& \textbf{92.0} & 94.0 & \textbf{100.0} & 94.0
& 82.0 & \textbf{94.0} & \textbf{90.0} & 96.0 & \textbf{100.0} \\

\bottomrule
\end{tabular}
\end{table*}

\begin{table}[t]
\centering
\caption{
Real-world manipulation performance on five tasks using an ARX X5 robot.
Each method is evaluated over 50 trials per task.
}
\label{tab:real_robot_results}
\small
\setlength{\tabcolsep}{5pt}

\begin{tabular}{l|ccccc|c}
\toprule
\multirow{2}{*}{\textbf{Method}}
& \multicolumn{6}{c}{\textbf{Success Rate (\%)}} \\
\cmidrule(lr){2-7}
& \shortstack{Pen}
& \shortstack{Color}
& \shortstack{Drawer}
& \shortstack{Towel}
& \shortstack{Flower}
& \textbf{Average} \\
\midrule

Diffusion Policy~\citep{chi2024diffusionpolicyvisuomotorpolicy}
& 78 & 56 & 78 & 80 & 52 & 68.8 \\

$\pi_{0.5}^*$~\citep{intelligence2025pi05visionlanguageactionmodelopenworld}
& \textbf{90} & 70 & 80 & \textbf{94} & 70 & 80.8 \\

BridgeVLA~\citep{li2025bridgevlainputoutputalignmentefficient}
& 88 & 70 & \textbf{86} & 74 & 68 & 77.2 \\

DeFI~\citep{ICLR2026_793bfa8f}
& 86 & 68 & 78 & 82 & 66 & 76.0 \\

\textbf{ReEnvision}
& \textbf{90} & \textbf{76} & 84 & 88 & \textbf{76} & \textbf{82.8} \\

\bottomrule
\end{tabular}
\end{table}

\begin{figure}[t]
    \centering
    \includegraphics[width=\linewidth]{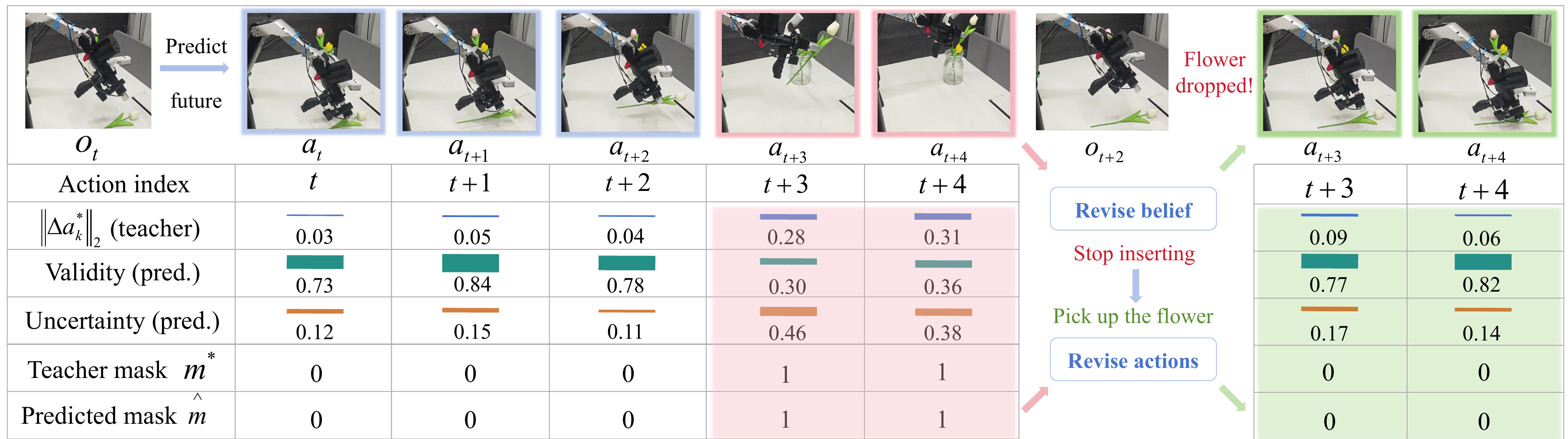}
    \caption{
    Visualization on a flower placement task. ReEnvision detects affected actions following an unexpected flower drop and selectively repairs the remaining action chunk through belief revision. 
For visualization, the teacher correction magnitudes \(\|\Delta a_k^*\|_2\) are rescaled for readability, with actions exceeding 0.10 on the rescaled scale marked as affected.}
    \label{fig:0907}
    \vspace{-1em}
\end{figure}

\section{Experiments}
\subsection{Experimental Setup}
We evaluate ReEnvision on CALVIN~\citep{mees2022calvinbenchmarklanguageconditionedpolicy}, L-CALVIN~\citep{fan2025long}, and RLBench~\citep{james2019rlbenchrobotlearningbenchmark}, using DeFI~\citep{ICLR2026_793bfa8f} as the backbone for CALVIN/L-CALVIN and BridgeVLA~\citep{li2025bridgevlainputoutputalignmentefficient} for RLBench. On CALVIN, we follow the ABC$\rightarrow$D protocol and evaluate 1,000 five-task sequences using 10-step, 7-DoF relative action chunks from RGB observations, proprioception, and language instructions. We report success rates for 1--5 consecutive tasks and average sequence length. L-CALVIN extends the same policy to ten-task sequences, reporting SR1--SR10 and average length. On RLBench, we evaluate 18 tasks with 50 episodes each and report per-task and mean success rates. All results are averaged over three random seeds.

For real-world evaluation, we use an ARX X5 robot on five tasks: pen placement, color sorting, sequential drawer placement, towel folding and placement, and sequential flower placement. We collect 150 kinesthetic demonstrations per task. Two Intel RealSense D405 cameras capture RGB observations at 30 Hz and $640\times480$ resolution. Each policy receives both RGB views, a language instruction, and a 7-D proprioceptive state.

\subsection{Ablation Studies}
We use DeFI and BridgeVLA as the baselines. For belief representation, the diffusion-policy conditioning is varied while the rest of ReEnvision is kept fixed: \(z_f\) only for DeFI, \(z_a\) only for action-only, and \(h_t=(z_f^t,z_a^t)\) for joint belief. Belief revision \(\Delta h\), uncertainty \(u_k\), and the full replanning are further ablated, with selective repair replaced by full replanning in the latter.

To isolate the effect of repair location, we fix the observation, task condition, and diffusion noise while changing the belief from \(h_{t+1}^{\mathrm{prior}}\) to \(h_{t+1}^{\mathrm{post}}\). Their paired policy responses define the reference correction \(\Delta A_{\mathrm{rem}}^{*}\), with actions above threshold \(\tau\) forming the reference mask \(m^{*}\). No Repair leaves the cached suffix unchanged without applying \(\Delta A_{\mathrm{rem}}^{*}\). For the remaining reference-mask controls, the same \(\Delta A_{\mathrm{rem}}^{*}\) is used while only the repair mask is varied: Selective w/ \(m^{*}\) repairs the affected positions, Random Mask repairs the same number of randomly selected positions, Complement Mask repairs positions outside \(m^{*}\), and Whole Suffix repairs the entire remaining suffix. 
ReEnvision applies the TCN-predicted mask \(\hat m\) together with the predicted correction \(\Delta\hat A_t^{\mathrm{rem}}\).

\begin{table*}[t]
\centering
\caption{
Ablation study of ReEnvision on RLBench, CALVIN, and L-CALVIN.
}
\label{tab:ablation}
\small
\setlength{\tabcolsep}{5pt}

\begin{tabular}{l|c|cc|ccc}
\toprule
\multirow{2}{*}{\textbf{Methods}}
& \textbf{RLBench}
& \multicolumn{2}{c|}{\textbf{CALVIN}}
& \multicolumn{3}{c}{\textbf{L-CALVIN}} \\
\cmidrule(lr){2-2}
\cmidrule(lr){3-4}
\cmidrule(lr){5-7}
& Avg. SR $\uparrow$
& 5-Step $\uparrow$
& Avg. Len. $\uparrow$
& Avg. Len. $\uparrow$
& 5-Step $\uparrow$
& 10-Step $\uparrow$ \\
\midrule

Baseline
& 85.7 & 81.2 & 4.51 & 7.30 & 0.75 & 0.44 \\

\midrule

$z_f$ only
& 84.6 & 84.7 & 4.58 & 8.22 & 0.85 & 0.55 \\

$z_a$ only
& 88.4 & 85.2 & 4.57 & 8.26 & 0.80 & 0.56 \\

w/o belief revision
& 87.4 & 82.6 & 4.54 & 7.96 & 0.83 & 0.50 \\

w/o uncertainty
& 88.0 & 85.2 & 4.60 & 8.32 & \textbf{0.90} & 0.61 \\

w/ full replanning
& \textbf{88.9} & 82.1 & 4.53 & 7.74 & 0.77 & 0.51 \\

\textbf{ReEnvision} $(z_f, z_a)$
& 88.8 & \textbf{86.6} & \textbf{4.62}
& \textbf{8.36} & 0.89 & \textbf{0.63} \\

\bottomrule
\end{tabular}
\end{table*}

\begin{table}[t]
\centering
\caption{
Repair-location control study on RLBench, CALVIN, and L-CALVIN.
}
\label{tab:repair_strategy}
\small
\setlength{\tabcolsep}{4.5pt}

\begin{tabular}{l|c|cc|ccc}
\toprule
\textbf{Methods}
& \textbf{RLBench}
& \multicolumn{2}{c|}{\textbf{CALVIN}}
& \multicolumn{3}{c}{\textbf{L-CALVIN}} \\
& Avg. SR $\uparrow$
& 5-Step $\uparrow$
& Avg. Len. $\uparrow$
& Avg. Len. $\uparrow$
& 5-Step $\uparrow$
& 10-Step $\uparrow$ \\
\midrule

No Repair
& 85.7 & 81.2 & 4.51 & 7.30 & 0.75 & 0.44 \\

Random Mask
& 83.1 & 81.6 & 4.48 & 7.59 & 0.80 & 0.49 \\

Complement Mask
& 77.4 & 78.8 & 4.39 & 6.85 & 0.69 & 0.38 \\

Selective w/ $m^*$
& \textbf{88.8} & 85.9 & \textbf{4.65} & \textbf{8.45} & \textbf{0.91} & \textbf{0.66} \\

Whole Suffix
& 81.2 & 80.7 & 4.50 & 7.15 & 0.80 & 0.43 \\

\textbf{ReEnvision}
& \textbf{88.8}  & \textbf{86.6} &  4.62 & 8.36 & 0.89 & 0.63 \\

\bottomrule
\end{tabular}
\end{table}

\subsection{Analysis}
As shown in Table~\ref{tab:calvin_results}, ReEnvision achieves the best performance from 3/5 onward and the highest Avg.~Len. of 4.62, with larger gains at longer horizons. This trend is clearer on L-CALVIN (Table~\ref{tab:lcalvin_results}), where ReEnvision reaches an Avg.~Len. of 8.36 and improves the later stages. On RLBench (Table~\ref{tab:rlbench_results}), it improves the reproduced BridgeVLA baseline from 85.7\% to 88.8\% average success rate. In real-world experiments (Table~\ref{tab:real_robot_results}), ReEnvision achieves the highest average success rate of 82.8\% across five manipulation tasks, with leading or tied-leading results on Pen, Color, and Flower. Overall, selective repair reduces error accumulation by preserving valid actions: 86.8\% of cached suffixes are kept unchanged, 11.1\% selectively patched, and only 2.1\% fully replanned.

Table~\ref{tab:ablation} shows consistent gains over the corresponding backbone baselines across all three benchmarks, especially in longer-horizon settings. Joint \((z_f,z_a)\) conditioning also outperforms either representation alone. Full replanning is slightly better on RLBench (88.9 vs.\ 88.8), while removing uncertainty gives a marginally higher L-CALVIN 5-step success (0.90 vs.\ 0.89); ReEnvision remains stronger on the longer-horizon metrics. In addition, Table~\ref{tab:repair_strategy} further highlights the importance of repair localization. On L-CALVIN, Selective w/ \(m^*\) reaches an Avg.~Len. of 8.45, compared with 7.59 for Random Mask and 6.85 for Complement Mask, showing that repairing the affected actions is more effective than modifying random or unaffected ones. Fig.~\ref{fig:0907} illustrates how ReEnvision selectively repairs affected actions following an unexpected flower drop.

\subsection{Discussion}
Our results suggest that belief changes usually affect only a local part of the cached action suffix rather than making the whole suffix invalid. The repair-location study further shows that repair location is critical: Selective w/ \(m^*\) outperforms Random Mask, Complement Mask, and Whole Suffix, supporting selective modification of affected actions while preserving valid ones. The ablations also show complementary roles for future prediction and action intent: the former captures expected task evolution, while the latter represents how the policy realizes it. Their combination therefore provides a stronger basis for deciding which remaining actions require revision. Full replanning remains competitive on RLBench, where tasks are shorter and rebuilding the suffix can correct errors with less disruption. In longer-horizon settings, repeated replanning may discard useful actions and introduce unnecessary changes, making ReEnvision increasingly advantageous as the execution horizon grows. The learned ReEnvision policy also performs close to Selective w/ \(m^*\), suggesting that these useful repair locations can be recovered without access to the reference mask.

ReEnvision estimates belief-induced action changes with a lightweight TCN and invokes full replanning only under severe deviations. Despite these optimizations, ReEnvision still introduces about 30\% additional inference latency over the baseline. In addition, the outcome-guided action pairs provide action-level supervision, but they do not establish whether each individual repair improves task success. Task-level success is difficult to use directly because it gives only a final outcome, making action-level credit assignment hard. Future work could first localize a small repair region using belief changes, then use task outcomes to refine which actions should be modified.

\section{Conclusion}
We presented ReEnvision, which combines the expected future and action intent into a joint belief, updates this belief with new observations, and estimates how the resulting revision affects the remaining action sequence. This enables the policy to preserve valid actions, selectively repair affected ones, and reserve full replanning for severe deviations. Experiments on CALVIN, L-CALVIN, and RLBench, together with real-robot evaluations, show consistent improvements over the corresponding backbone policies, with larger gains over longer horizons. The repair-location study further shows that performance gains depend on correctly localizing belief-induced changes within the action chunk, so that affected actions are corrected while valid actions are preserved. Future work may improve affected-action estimation and extend selective repair to more diverse real-world settings.

\section*{AI Use Statement}
In this work, generative AI tools were used to assist with literature search and synthesis and to improve the clarity, readability, and wording of the manuscript. All AI-assisted content, including literature-related information and textual revisions, was reviewed and verified by the authors against the original sources where applicable. The authors take full responsibility for the final content, claims, and presentation of this work.

\section*{Reproducibility Statement}
We provide the model architecture, training objectives, and staged training and inference procedures in Section 3. Section 4 details the benchmark protocols, evaluation metrics, and real-world experimental setup. Appendix A includes additional experimental analyses, runtime measurements, and hyperparameter sensitivity studies. All source code and pretrained model weights will be publicly released upon acceptance to facilitate reproducibility.

\bibliography{iclr2027_conference}
\bibliographystyle{iclr2027_conference}

\appendix

\section{Supplementary experiments}
\subsection{Mechanism Validation}
\label{app:mechanism}
We further validate two properties underlying selective action repair:
(i) belief revision typically affects only a localized portion of the cached
action suffix, and
(ii) the TCN can accurately localize and predict the corresponding action changes.

\paragraph{Locality of affected actions.}
We use the paired policy responses
under $h_{t+1}^{\mathrm{prior}}$ and $h_{t+1}^{\mathrm{post}}$ to obtain
$\Delta A_{\mathrm{rem}}^{*}$ and the corresponding reference affected-action
mask $m^{*}$. Specifically, $m_k^{*}=1$ indicates that the change of action
$k$ exceeds the threshold $\tau$. We measure the fraction of affected actions
and the number of contiguous affected segments within each cached suffix.

\paragraph{TCN localization accuracy.}
The TCN-predicted affected-action mask is evaluated against $m^{*}$.
We report mask F1, temporal and segment IoU, and the prediction error of the action correction \(\Delta\hat A_t^{\mathrm{rem}}\).

\begin{table}[t]
\centering
\caption{
Validation of action-change locality and TCN localization accuracy.
$m^{*}$ denotes the reference affected-action mask derived from paired policy
responses. Non-empty IoU is computed only on suffixes containing at least one
affected action.
}
\label{tab:mechanism_validation}
\small
\setlength{\tabcolsep}{5pt}
\begin{tabular}{l|c}
\toprule
\textbf{Metric} & \textbf{Value} \\
\midrule
Affected fraction ($m^{*}$) & 0.208 \\
Pred. affected fraction & 0.206 \\
Segments per suffix ($m^{*}$) & 0.821 \\
Pred. segments per suffix & 0.850 \\
\midrule
Mask F1 & 0.861 \\
Temporal IoU (all) & 0.785 \\
Temporal IoU (non-empty) & 0.690 \\
Segment IoU (all) & 0.788 \\
Segment IoU (non-empty) & 0.696 \\
$\Delta A$ MAE & 0.000167 \\
\bottomrule
\end{tabular}
\end{table}

As shown in Table~\ref{tab:mechanism_validation}, only 20.8\% of the cached suffix is marked as affected by $m^{*}$,
with 0.821 contiguous affected segments per suffix, indicating that
belief-induced action changes are typically localized rather than distributed
throughout the suffix. The TCN closely reproduces this structure, achieving a
mask F1 of 0.861 and a non-empty segment IoU of 0.696 while predicting a similar
affected fraction (20.6\%). 
These results demonstrate that belief-induced action changes can be reliably localized. Notably, predicted action impact does not necessarily lead to modification, as the final intervention is determined by structured mask selection and the keep-repair-replan decision.

\subsection{Controlled Deviation Recovery}
We construct a controlled deviation evaluation on L-CALVIN to examine how different recovery strategies respond to execution errors of increasing severity. During rollout, deviations are injected directly into the simulator state using two complementary perturbations. First, an end-effector deviation applies a controlled offset to the robot pose, providing a simple and reproducible execution error whose magnitude can be precisely varied. Second, object displacement perturbs the pose of a task-relevant object, representing an external state change that may invalidate only part of the cached action suffix. Both perturbations are applied at small, medium, and severe levels. This setup allows us to test whether localized deviations can be handled by selectively modifying the affected actions, while larger deviations increasingly require replanning. Recovery Success denotes the fraction of perturbed rollouts that successfully recover, Avg. Len. measures the average number of consecutively completed tasks, Modified Actions denotes the fraction of the cached suffix modified during recovery, and Replan Rate reports the frequency of discarding the cached suffix and generating a new action chunk. The clean setting corresponds to ReEnvision without injected perturbations and serves as a reference for the performance degradation caused by execution deviations.
\begin{table*}[t]
\caption{
Controlled deviation recovery on L-CALVIN under perturbations of increasing severity.
}
\label{tab:controlled_deviation}
\centering
\small
\setlength{\tabcolsep}{6pt}
\begin{tabular}{llcccc}
\toprule
Severity & Strategy
& Recovery Success $\uparrow$
& Avg. Len. $\uparrow$
& Modified Actions
& Replan Rate \\
\midrule

Clean
& ReEnvision
& 0.75 & 8.36 & 0.08 & 0.03 \\

\midrule

Small
& No Repair
& 0.79 & 7.18 & 0.00 & 0.00 \\

Small
& Step-wise
& 0.75 & 7.56 & 0.03 & 0.00 \\

Small
& Full Replan
& 0.68 & 7.07 & 1.00 & 1.00 \\

Small
& Oracle Selective
& 0.93 & 8.25 & 0.08 & 0.05 \\

Small
& ReEnvision
& 0.84 & 7.93 & 0.18 & 0.11 \\

\midrule

Medium
& No Repair
& 0.55 & 6.96 & 0.00 & 0.00 \\

Medium
& Step-wise
& 0.55 & 7.10 & 0.06 & 0.00 \\

Medium
& Full Replan
& 0.63 & 6.95 & 1.00 & 1.00 \\

Medium
& Oracle Selective
& 0.78 & 7.30 & 0.20 & 0.13 \\

Medium
& ReEnvision
& 0.70 & 7.14 & 0.30 & 0.17 \\

\midrule

Severe
& No Repair
& 0.35 & 5.68 & 0.00 & 0.00 \\

Severe
& Step-wise
& 0.42 & 5.75 & 0.08 & 0.00 \\

Severe
& Full Replan
& 0.58 & 6.21 & 1.00 & 1.00 \\

Severe
& Oracle Selective
& 0.62 & 6.91 & 0.36 & 0.38 \\

Severe
& ReEnvision
& 0.60 & 6.62 & 0.37 & 0.52 \\

\bottomrule
\end{tabular}
\end{table*}

\begin{figure*}[t]
    \centering
    \includegraphics[width=0.88\textwidth]{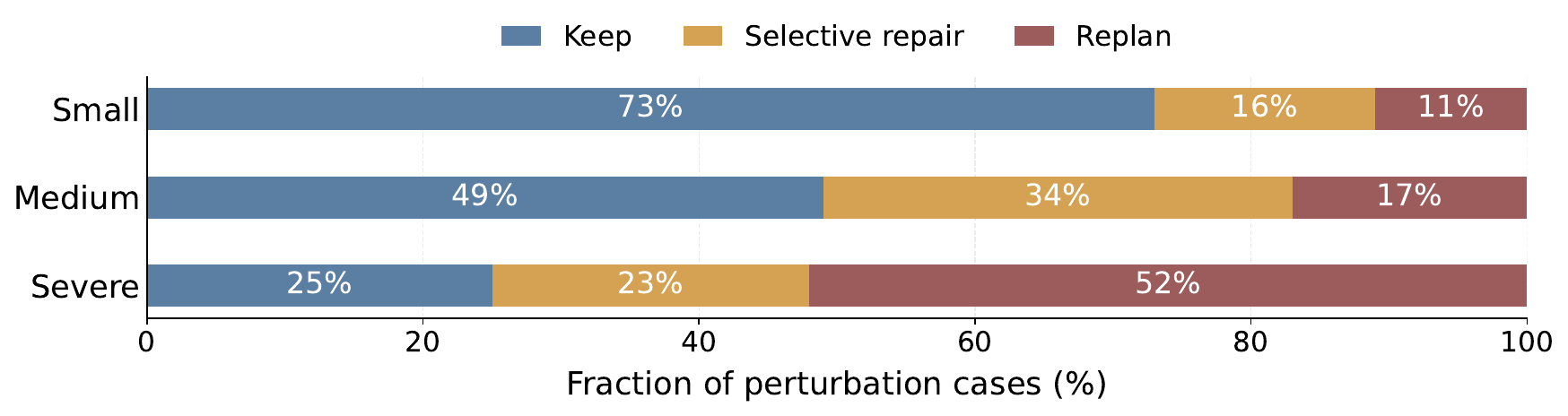}
    \caption{
    ReEnvision's execution strategy under controlled deviations of increasing severity.
    Each bar shows the fraction of perturbation cases handled by keeping the cached suffix,
    selectively repairing it, or triggering full replanning.
    As deviation severity increases, ReEnvision progressively shifts from suffix preservation
    toward selective intervention and full replanning.
    }
    \label{fig:strategy_shift}
\end{figure*}

As shown in Table~\ref{tab:controlled_deviation}, recovery becomes more difficult as deviation severity increases, while ReEnvision progressively shifts from suffix reuse and selective repair toward more frequent replanning. Interestingly, under small deviations, No Repair performs better than both Step-wise correction and Full Replan, suggesting that weak perturbations often leave much of the cached suffix valid and that unnecessary intervention can disrupt otherwise successful execution. This also explains why selective repair provides the largest benefit when the deviation is large enough to invalidate part of the suffix but not severe enough to require complete regeneration. Under severe deviations, ReEnvision approaches the recovery performance of Full Replan while maintaining a higher Avg. Len., indicating that even in this regime, some cached actions can remain useful. Overall, these results support adapting the scope of intervention to the extent of the execution deviation.

Figure~\ref{fig:strategy_shift} further illustrates how ReEnvision adapts its
execution strategy to deviation severity. Under small deviations, most cases are
handled by preserving the cached suffix, with a keep rate of 0.73. As the deviation
becomes moderate, selective repair becomes more frequent, increasing from 0.16 to
0.34, while the keep rate decreases. Under severe deviations, full replanning becomes
the dominant response, accounting for 0.52 of the cases. This progression shows that
ReEnvision adjusts the scope of intervention according to the extent of the execution
deviation, preserving existing actions when possible and increasingly regenerating
the suffix when the cached plan becomes unreliable.

\subsection{Efficiency and Runtime Analysis}

\begin{table}[t]
\centering
\caption{
Runtime analysis of ReEnvision. Component latencies are measured individually, while the total latency reports end-to-end inference time per execution step. ReEnvision introduces approximately 32\% additional latency over the backbone policy, while the belief revision and selective repair components add only a small computational overhead.
}
\label{tab:runtime}
\small
\setlength{\tabcolsep}{5pt}

\begin{tabular}{lc}
\toprule
\textbf{Component} & \textbf{Latency} \\
\midrule
Backbone policy & $\sim$110 ms \\
Future--action belief & $\sim$8 ms \\
Belief revision model & $\sim$2 ms \\
Temporal impact model & $\sim$1 ms \\
Mask optimization & $<$1 ms \\
\midrule
\textbf{ReEnvision (end-to-end)} & $\sim$145 ms \\
Full diffusion replan & $\sim$110 ms \\
\bottomrule
\end{tabular}
\end{table}

Table~\ref{tab:runtime} reports the inference cost of the main ReEnvision components. The backbone diffusion policy requires approximately 110~ms per inference, while belief construction, RSSM updating, temporal impact prediction, and mask optimization are comparatively lightweight. In particular, the temporal impact model requires only about 1~ms, allowing belief-induced action changes to be estimated without an additional diffusion-policy evaluation. The complete ReEnvision pipeline requires approximately 145~ms per execution step, corresponding to about 32\% additional end-to-end latency over the backbone. Full diffusion replanning, in contrast, requires an additional policy evaluation of approximately 110~ms when triggered. Since ReEnvision invokes full replanning only for severe deviations, most execution steps are handled through substantially cheaper keep or selective-repair operations.

\subsection{Hyperparameter Sensitivity}

We examine the sensitivity of ReEnvision to three parameters: the affected-action threshold \(\tau\), the maximum repair ratio \(\rho\), and the uncertainty threshold \(\eta\). The maximum number of repair segments is fixed at \(s_{\max}=1\). While \(\tau\) controls offline supervision of the action impact model, \(\rho\) and \(\eta\) govern the online keep-repair-replan decision.

During training, \(\tau\) defines the outcome-guided affected-action labels \(m_k^*=\mathbb{I}[\|\Delta a_k^*\|_2>\tau]\), the corresponding validity targets \(v_k^*=1-m_k^*\), and the uncertainty targets. At inference, \(S_\psi\) predicts action corrections, validity, and uncertainty, which are used for structured mask selection without directly thresholding the predicted corrections using \(\tau\). The maximum uncertainty \(u_{\max}\) is computed across the entire cached suffix. Full replanning is triggered when \(u_{\max}\geq\eta\); otherwise, an empty mask retains the suffix, while a nonempty mask is selectively repaired when \(r\leq\rho\) and \(s\leq s_{\max}\). If either condition is violated, the suffix is fully replanned.

Sensitivity is evaluated on CALVIN using the same 5-step evaluation protocol as the main experiments. The default configuration is \(\tau=10^{-4}\), \(\rho=0.6\), \(s_{\max}=1\), and \(\eta=0.9\). For each value of \(\tau\), the validity and uncertainty targets are regenerated, and \(S_\psi\) is retrained using the same training protocol. All variants are evaluated using the same structured mask selection rule, with \(\rho\) and \(\eta\) fixed at their default values. To evaluate \(\rho\) and \(\eta\), each inference threshold is varied independently while the action impact model trained with the default \(\tau\) and all other parameters remain fixed. Specifically, we evaluate \(\tau\in\{5\times10^{-5},10^{-4},2\times10^{-4}\}\), \(\rho\in\{0.4,0.6,0.8\}\), and \(\eta\in\{0.7,0.9,1.0\}\).

\begin{table}[t]
\centering
\caption{
Hyperparameter sensitivity on CALVIN. Each parameter is varied independently while the others are fixed at their default values: $\tau=10^{-4}$, $\rho=0.6$, and $\eta=0.9$. The default configuration is shown in bold.
}
\label{tab:hyperparameter_sensitivity}
\small
\setlength{\tabcolsep}{5pt}
\begin{tabular}{ccccc}
\toprule
Parameter & Value & Avg. Len. $\uparrow$ & 5-Step SR (\%) $\uparrow$ & Replan Rate (\%) \\
\midrule
\multirow{3}{*}{$\tau$}
& $5\times10^{-5}$ & 4.50 & 85.0 & 8.0 \\
& $\mathbf{1\times10^{-4}}$ & \textbf{4.62} & \textbf{86.6} & \textbf{2.1} \\
& $2\times10^{-4}$ & 4.55 & 82.0 & 3.0 \\
\midrule
\multirow{3}{*}{$\rho$}
& 0.4 & 4.59 & 87.0 & 12.0 \\
& $\mathbf{0.6}$ & \textbf{4.62} & \textbf{86.6} & \textbf{2.1} \\
& 0.8 & 4.57 & 86.0 & 4.0 \\
\midrule
\multirow{3}{*}{$\eta$}
& 0.7 & 4.53 & 83.0 & 15.0 \\
& $\mathbf{0.9}$ & \textbf{4.62} & \textbf{86.6} & \textbf{2.1} \\
& 1.0 & 4.53 & 85.0 & 3.0 \\
\bottomrule
\end{tabular}
\end{table}

As shown in Table~\ref{tab:hyperparameter_sensitivity}, the default configuration achieves the highest average sequence length among the evaluated settings. Varying \(\tau\) changes the offline supervision of \(S_\psi\) and consequently its predictions during execution. A smaller \(\tau\) increases the replan rate to 8.0\%, while a larger \(\tau\) reduces the 5-step success rate to 82.0\%. For the online decision thresholds, a stricter repair ratio of \(\rho=0.4\) increases the replan rate to 12.0\%, with a slightly lower average sequence length despite a marginally higher 5-step success rate. Similarly, reducing \(\eta\) to 0.7 increases the replan rate to 15.0\% and decreases the average sequence length to 4.53. These results illustrate how offline affected-action supervision and online intervention thresholds jointly influence repair behavior, with the default configuration achieving the highest average sequence length in this study.

\section{Theoretical Analysis}
\subsection{Uncertainty Target}
For each future action, we define the magnitude of the outcome-guided correction as
\begin{equation}
s_k^* = \|\Delta a_k^*\|_2,
\label{eq:unc_action_change}
\end{equation}
and determine whether the action is affected using
\begin{equation}
m_k^* = \mathbb{I}[s_k^* > \tau],
\label{eq:unc_affected_label}
\end{equation}
where $\tau > 0$ is the affected-action threshold.

Since predictions near the decision boundary $s_k^* = \tau$ are more ambiguous, we define the uncertainty target as
\begin{equation}
u_k^*
=
\exp\left(
-\frac{|s_k^*-\tau|}{\tau}
\right).
\label{eq:uncertainty_target}
\end{equation}

\paragraph{Proposition.}
The uncertainty target $u_k^*$ reaches its maximum at $s_k^*=\tau$ and decreases monotonically as $s_k^*$ moves away from the decision boundary.

\paragraph{Proof.}
Let
\begin{equation}
d_k = |s_k^*-\tau|
\label{eq:unc_distance}
\end{equation}
denote the distance from the decision boundary. Equation~\ref{eq:uncertainty_target} can then be rewritten as
\begin{equation}
u_k^* = \exp\left(-\frac{d_k}{\tau}\right).
\label{eq:uncertainty_distance}
\end{equation}
Because $\tau > 0$,
\begin{equation}
\frac{\partial u_k^*}{\partial d_k}
=
-\frac{1}{\tau}
\exp\left(-\frac{d_k}{\tau}\right)
< 0.
\label{eq:uncertainty_derivative}
\end{equation}
Therefore, $u_k^*$ decreases monotonically with the distance $d_k$. When $s_k^*=\tau$, we have $d_k=0$, yielding $u_k^*=1$, which is the maximum value. Hence, actions closer to the affected-action decision boundary receive higher uncertainty targets, while actions farther from the boundary receive lower uncertainty targets. The uncertainty target therefore provides a monotonic measure of ambiguity around the affected-action decision boundary.

\subsection{Controlled Belief Intervention}

To isolate the effect of belief revision, we generate paired action chunks \(A^{-}\) and \(A^{+}\) using the same frozen policy. The observation, task, diffusion noise, and cache state are held fixed, while only the belief changes from \(h^{-}\) to \(h^{+}\). Their first \(K-1\) actions, denoted by \(A_{\mathrm{rem}}^{-}\) and \(A_{\mathrm{rem}}^{+}\), are temporally aligned with the cached suffix. For this analysis, the reference correction is defined as
\begin{equation}
\Delta A_{\mathrm{rem}}^*
= A_{\mathrm{rem}}^+ - A_{\mathrm{rem}}^-.
\label{eq:belief_intervention_difference}
\end{equation}

\paragraph{Proposition.}
Suppose the paired actions are generated by the same frozen policy under identical observation, task, diffusion sampling condition, and cache state, with only the belief condition changing from \(h^-\) to \(h^+\). Then, \(\Delta A_{\mathrm{rem}}^*\) isolates the policy's action response to the belief change \(\Delta h=h^+-h^-\). 
If the policy has a locally Lipschitz-continuous Jacobian with respect to its belief condition, this response admits a first-order approximation with a second-order remainder.

\paragraph{Proof.}
With all other policy inputs fixed, let \(P_{\mathrm{rem}}\) select the first \(K-1\) actions and define
\begin{equation}
F_{\mathrm{rem}}(h)
= P_{\mathrm{rem}}\left(\pi(o_{t+1},l,h;\xi)\right).
\label{eq:fixed_policy_function}
\end{equation}

Under the stated matching conditions, the paired action difference satisfies
\begin{equation}
\Delta A_{\mathrm{rem}}^*
= F_{\mathrm{rem}}(h^+)-F_{\mathrm{rem}}(h^-).
\label{eq:belief_only_difference}
\end{equation}

Assuming a locally Lipschitz-continuous Jacobian, a first-order Taylor expansion gives
\begin{equation}
\Delta A_{\mathrm{rem}}^*
= J_hF_{\mathrm{rem}}(h^-)\Delta h
+ \mathcal{O}\left(\|\Delta h\|_2^2\right).
\label{eq:counterfactual_first_order}
\end{equation}

For sufficiently small belief changes,
\begin{equation}
\Delta A_{\mathrm{rem}}^*
\approx J_hF_{\mathrm{rem}}(h^-)\Delta h.
\label{eq:counterfactual_approx}
\end{equation}

Thus, under the matching conditions, the paired target captures the frozen policy's sensitivity to belief changes. 
The temporal impact model \(S_\psi\) is trained on outcome-guided pairs to approximate belief-induced action changes without explicitly computing the policy Jacobian. During inference, pre-revision actions are retrieved directly from the cached suffix \(A_t^{\mathrm{rem}}\), and predicted corrections are applied only to selected actions without regenerating the pre-revision chunk.

\subsection{Property of Selective Repair}

Selective repair updates the cached action suffix according to
\begin{equation}
A_t^{\mathrm{rem}\prime}=A_t^{\mathrm{rem}}+\hat{\mathbf{m}}\odot\Delta\hat{A}_t^{\mathrm{rem}},
\label{eq:selective_repair_app}
\end{equation}
where \(\hat{\mathbf{m}}\in\{0,1\}^{K-1}\) denotes the selected repair mask, and \(\Delta\hat{A}_t^{\mathrm{rem}}\) denotes the predicted correction for the cached suffix. Each \(k=1,\ldots,K-1\) indexes the corresponding future action \(a_{t+k}\).

\paragraph{Proposition.}
Selective repair preserves all unselected actions and introduces no larger action perturbation than applying the same correction to the entire suffix. Specifically, for every \(k\) with \(\hat{m}_k=0\),
\begin{equation}
a_{t+k}'=a_{t+k},
\label{eq:unaffected_preserved}
\end{equation}
and, for any \(p\geq1\),
\begin{equation}
\left\|A_t^{\mathrm{rem}\prime}-A_t^{\mathrm{rem}}\right\|_p
=
\left\|\hat{\mathbf{m}}\odot\Delta\hat{A}_t^{\mathrm{rem}}\right\|_p
\leq
\left\|\Delta\hat{A}_t^{\mathrm{rem}}\right\|_p.
\label{eq:selective_bound}
\end{equation}

\paragraph{Proof.}
From Eq.~\ref{eq:selective_repair_app}, each remaining action is updated as
\begin{equation}
a_{t+k}'=a_{t+k}+\hat{m}_k\Delta\hat{a}_k.
\label{eq:actionwise_repair_app}
\end{equation}
When \(\hat{m}_k=0\), the correction vanishes and \(a_{t+k}'=a_{t+k}\), so every unselected action is preserved.

Moreover, since \(\hat{m}_k\in\{0,1\}\), masking cannot increase the magnitude of the applied correction. Therefore,
\begin{equation}
\left\|\hat{\mathbf{m}}\odot\Delta\hat{A}_t^{\mathrm{rem}}\right\|_p^p
=
\sum_{k=1}^{K-1}\hat{m}_k\left\|\Delta\hat{a}_k\right\|_p^p
\leq
\sum_{k=1}^{K-1}\left\|\Delta\hat{a}_k\right\|_p^p
=
\left\|\Delta\hat{A}_t^{\mathrm{rem}}\right\|_p^p.
\label{eq:masked_norm_bound}
\end{equation}
Taking the \(p\)-th root gives Eq.~\ref{eq:selective_bound}. Thus, selective repair preserves unselected actions while restricting the action perturbation to the selected subset.

\end{document}